\documentclass[journal]{IEEEtran}
\usepackage[noadjust]{cite}
\usepackage[cmex10]{amsmath}
\usepackage[caption=false,font=footnotesize]{subfig}
\usepackage{url}
\usepackage{graphicx}
\usepackage{makecell}
\usepackage{multirow}
\usepackage{todonotes}

\newcolumntype{L}[1]{>{\raggedright\let\newline\\\arraybackslash\hspace{0pt}}m{#1}}
\newcolumntype{C}[1]{>{\centering\let\newline\\\arraybackslash\hspace{0pt}}m{#1}}
\newcolumntype{R}[1]{>{\raggedleft\let\newline\\\arraybackslash\hspace{0pt}}m{#1}}

\begin{document}
%\title{IEEE Journal Template Example}
%\title{Provenance Analysis for Digital Images:\\An End-to-End Approach}
%\title{Image Provenance Analysis at Web Scale}
\title{Image Provenance Analysis at Scale}

\author{Daniel~Moreira, %~\IEEEmembership{Member,~IEEE,}
        Aparna~Bharati,~\IEEEmembership{Student~Member,~IEEE,}
        Joel~Brogan,~\IEEEmembership{Student~Member,~IEEE,}
        Allan~Pinto,~\IEEEmembership{Student~Member,~IEEE,}
        Michael~Parowski,
        Kevin~W.~Bowyer,~\IEEEmembership{Fellow,~IEEE,}
        Patrick~J.~Flynn,~\IEEEmembership{Fellow,~IEEE,}
        Anderson~Rocha,~\IEEEmembership{Senior~Member,~IEEE,}
        and~Walter~J.~Scheirer,~\IEEEmembership{Senior~Member,~IEEE}% <-this % stops a space
\thanks{D. Moreira, A. Bharati, J. Brogan, M. Parowski, K. Bowyer, P. Flynn, and W. Scheirer are with the Department of Computer Science and Engineering, University of Notre Dame, IN, 46556, USA.}% %Corresponding author's e-mail: walter.scheirer@nd.edu.}% <-this % stops a space
\thanks{A. Pinto and A. Rocha are with the Institute of Computing, University of Campinas, SP, 13083-852, Brazil.} 
\thanks{Corresponding author: W. J. Scheirer (walter.scheirer@nd.edu).}}
\maketitle

\begin{abstract}
Prior art has shown it is possible to estimate, through image processing and computer vision techniques, the types and parameters of transformations that have been applied to the content of individual images to obtain new images.
Given a large corpus of images and a query image, an interesting further step is to retrieve the set of original images whose content is present in the query image, as well as the detailed sequences of transformations that yield the query image given the original images.
This is a problem that recently has received the name of image provenance analysis.
% Why do we care?
In~these times of public media manipulation ({\em e.g.}, fake news and meme sharing), obtaining the history of image transformations is relevant for fact checking and authorship verification, among many other applications.
% How do we solve?
This article presents an end-to-end processing pipeline for image provenance analysis, which works at real-world scale.
It employs a cutting-edge image filtering solution that is custom-tailored for the problem at hand, as well as novel techniques for obtaining the provenance graph that expresses how the images, as nodes, are ancestrally connected.
% What are the relevant contributions and results?
A comprehensive set of experiments for each stage of the pipeline is provided, comparing the proposed solution with state-of-the-art results, employing previously published datasets.
In addition, this work introduces a new dataset of real-world provenance cases from the social media site \textit{Reddit}, along with baseline results. % obtained by the proposed solution.

\end{abstract}

\begin{IEEEkeywords}
%IEEEtran, journal, \LaTeX, paper, template.
Digital Image Forensics, Digital Humanities, Image Retrieval, Graphs, Image Provenance, Image Phylogeny
\end{IEEEkeywords}

% \fi
\IEEEpeerreviewmaketitle

\section{Introduction}
\label{sec:intro}

Algorithms for the detection of manipulated content in digital images have reached a stage of maturity that is sufficient for understanding the transformations that were applied to individual images in many cases~\cite{farid2009image,Rocha:CSUR:2011,farid2017detect}.
A logical next step is to develop an approach that allows us to ask more complicated questions about the relationships between related images after sequences of transformations have been applied --- a problem that is not well studied in the image processing literature.
In this article, we consider the \textit{Provenance Analysis} task~\cite{Dias_2012,Dias_2013}, in which the objective is to recover the graph of relationships between plausibly connected images. These relationships may be expressed as undirected edges (\textit{i.e.}, neighboring transformations are identified) or directed edges (\textit{i.e.}, the order of neighboring transformations is expressed).
The development of techniques to recover such graphs combines ideas from the areas of image retrieval, digital image forensics, and graph theory, making this an interesting interdisciplinary endeavour within image processing and computer vision. 

%(Teaser Figure; meme example)
\begin{figure}[t]
\centering
\vspace{-0.3cm}
\includegraphics[width=8.7cm]{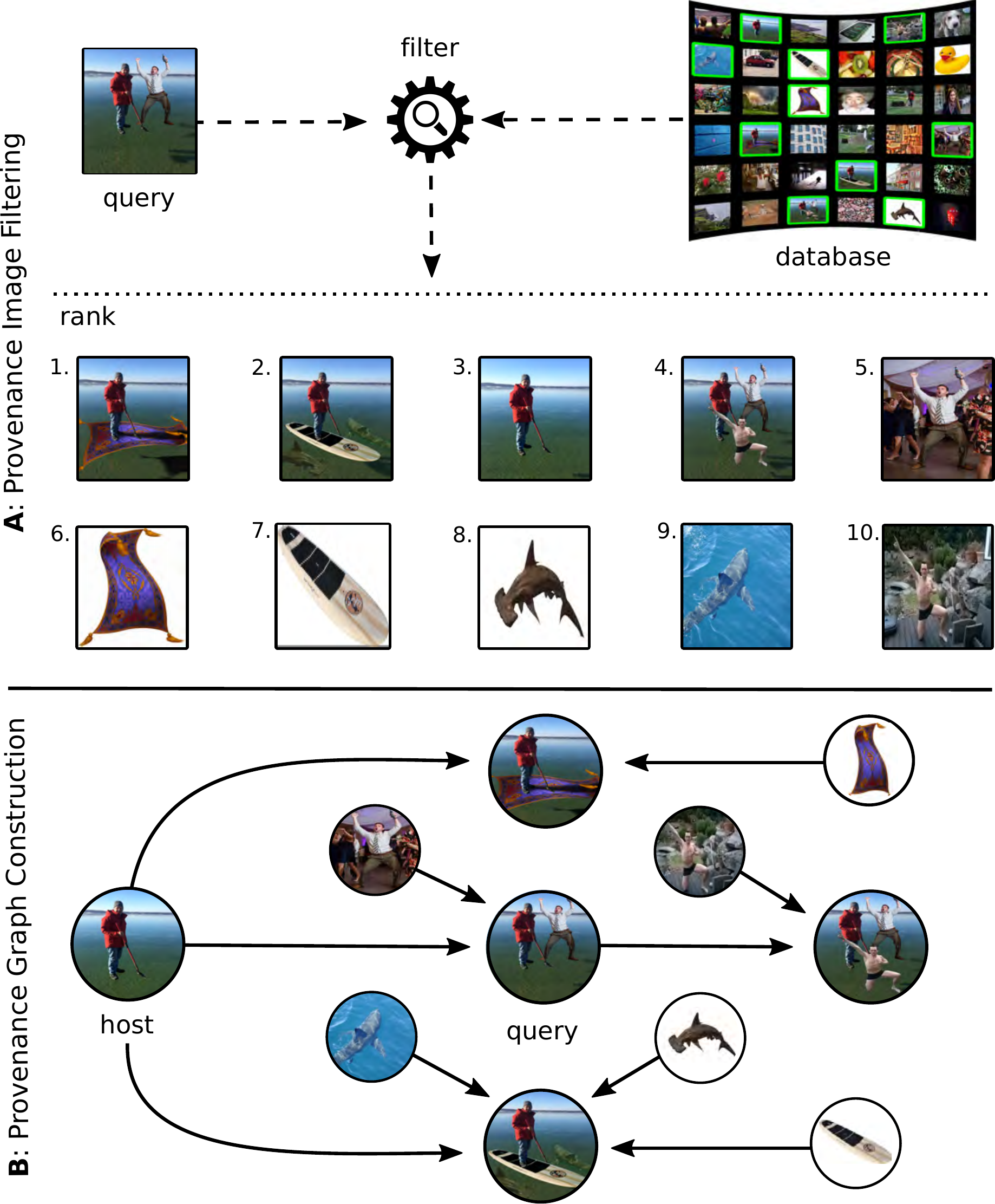}
\caption[Image Provenance Analysis]{Image Provenance Analysis workflow.
Panel A depicts the first step of Image Provenance Analysis, namely Provenance Image Filtering, in which filters are applied to a large image database to retrieve those images that are related to a given query image.
Panel B depicts the second step, namely Provenance Graph Construction, in which the filtered images are linked to each other in a way that expresses the sequences of manipulation and/or compositions (\textit{i.e.}, the provenance history of the images).}
%\vspace{-0.2cm}
\label{fig:teaser}
\end{figure}

%In more concrete terms, what is the provenance analysis task?
To illustrate the provenance analysis task, consider the set of example images in Panel A of Fig.~\ref{fig:teaser}, which were collected from the popular ``Photoshop battles'' forum on the social media site Reddit~\cite{reddit2017photoshopbattles}. On this forum, amateur artists begin with source images and employ image manipulation tools to generate results for humorous effect.
The first step in provenance analysis is \textit{Provenance Image Filtering}, which consists of searching a potentially large pool of images for those that are most closely related to a given query image.
Related images might be {\em semantically similar} ({\em i.e.}, the same scene may be present from slightly different view points or at nearby points in time), or they might be {\em near duplicates} related by minor transformations such as exposure and saturation adjustments, or cropping and re-sizing, or they might be {\em  image compositions}, which contain elements of two or more different source images. In most cases, the query will be an image that has been manipulated in some way. 

The second step is \textit{Provenance Graph Construction}, where the objective is to understand the relationships between images yielded by provenance image filtering.
A \textit{Host Image} provides the source of background content for subsequent manipulations.
In Fig.~\ref{fig:teaser}, the host is the photo of the man holding a shovel in the 
leftmost part of Panel B. % (itself a likely copy-move forgery~\cite{fridrich2003detection}).
A \textit{Donor Image} provides some amount of content that will be inserted into a host image.
In Fig.~\ref{fig:teaser}, three donor images are the original images of the sharks and the paddle board in the %center 
bottom half of Panel B.
They provide image content that has been inserted into the image %to their left
they are linked to.
Sequences of manipulations are common, and they can be expressed as a directed graph representing the order in which they were applied.
This can be seen in the %overall graph in Panel B,
graph of Panel B, where the depth %of certain paths
of the central path containing the host and the query 
leads to three different levels of manipulations.
Our goal %in this work
is to develop an algorithm that can generate such graphs in an automated fashion.
We do not make strong assumptions that either the original host or donor images are available during analysis. %For instance, the surfboard, magic carpet, elephant, and extra people were not harvested at the image retrieval stage of the process.
For instance, the paddle board, flying carpet, and extra people might not necessarily be harvested at the image filtering step.
% In addition, the analysis does not have any restriction on the number and type of unrelated images (distractors) to be assessed in a potential pool of images

%Why is provenance analysis important to image processing?
Provenance analysis is important to image processing and computer vision.
%This problem is not posed as just a thought experiment.
%On the contrary,
It has direct applications in a number of different fields.
The most immediate application is forensics, where the detection of manipulated images spans traditional policing to analysis for strategic intelligence.
The question of the origins of suspect images has taken a prominent role recently, with the rise of so-called ``fake news" on the Internet. 
While not a new problem\footnote{The computer hacker group Cult of the Dead Cow warned of the devastating potential of widespread online media manipulation as early as 1999~\cite{cdc}.}, concern about fake news reached new heights on the heels of the 2016 American presidential election. The rapid evolution of the online social media landscape has provided new, free media channels with which even amateur bloggers and news outlets can reach massive audiences with little effort, and even less regulation. Recent instances of fake news often involve questionable images propagating through social media.
For example, in early 2017, the New York Times reported on the creation of a false story about the discovery of pre-marked ballots in Ohio that appeared a couple of months before the election~\cite{NYT1}.
The image accompanying the story was the product of a mirrored image that was selectively blacked-out in local regions~\cite{PP1}.
This is a real-life case with multiple manipulations where provenance analysis could be applied to trace the origins of the fabrication. 
% Expert analysis cannot feasibly be deployed to the massive influx of data that bombards average internet users on a daily basis, such as the article above. Unfortunately, users tend resonate with stories and media that appeal to them emotionally and viscerally \cite{pronin2007valuing}. 

%% Probably out of scope for this paper
%%A related application is plagiarism / scientific %misconduct detection. (cite) Where  the content came from %can shed some light on the case.

Beyond the important application domain of forensics, image provenance analysis can form a powerful framework for academic research in other fields.
Cultural analytics has emerged as a distinct sub-discipline within the digital humanities~\cite{manovich2009cultural,yamaoka2011cultural} that is concerned with combining quantitative methods from social science and computer science to answer humanistic questions about cultural trends.
An example of this (which we have already touched upon in Fig.~\ref{fig:teaser}) is the study of Internet \textit{memes} --- cultural artifacts meant to be widely transmitted and evolve over time.
Memes are an interesting object of cultural study, in that they encapsulate facets of popular entertainment, political moods, and novel elements of humor.
Meme aggregators like the website \emph{knowyourmeme.com} have done a good job at archiving such content, but a more exhaustive quantitative study of the provenance of individual memes has yet to emerge.
Tracing the source(s) of modified meme images helps us unpack the underlying cultural trends that can tell us something meaningful about the community that generated the content.

%% What algorithmic components are necessary to solve this problem?
%% First, one needs an accurate and scalable image retrieval algorithm that is able to operate over very large collections of images (realistically, on the order of millions of images) to find related candidates.
%% Second, identification of image transformations and the localization of any tampering provide evidence for the ordering of the related images.
%%And third, methods from graph theory are necessary to establish the relationships between images, yielding a directed graph that is interpretable by a human analyst.
%% Further, all of these components must be integrated to solve the problem as a coherent processing pipeline.
%What algorithmic components are necessary to solve this problem?
Both of the application domains mentioned also motivate the need for any developed techniques to be {\em scalable}. Specialized algorithmic components are necessary to solve the problem at hand.
First, one needs an accurate and scalable image retrieval algorithm that is able to operate over very large collections of images (realistically, on the order of millions of images) to find related candidates.
Such an algorithm also has to address the particularities of the provenance image filtering task: it must perform well at retrieving the near-duplicate host images that are highly related to the query (a well-known problem in the image retrieval literature), but also perform well at retrieving donors (images that potentially donated small portions to the query) and the donors' respective near duplicates (which might not be directly related to the query).
Second, the identification of likely image transformations that explain how each retrieved image might have been used to generate the others is required, as it is used to create the ordering of the images in the provenance graph.
And third, methods from graph theory are necessary to organize the relationships between images, yielding a directed graph that is human-interpretable. % by a human analyst.
All of these components must be integrated %to solve the problem
as a coherent and scalable processing pipeline.

This work introduces, for the first time, a fully automated large-scale end-to-end pipeline that starts with the step of provenance image filtering (over millions of images) and ends up with the provenance graphs.
The following new contributions are introduced this work:

\vspace{0.2cm}
\begin{enumerate}
\item \emph{Distributed interest point selection}: a novel interest point selection strategy that aims at spatially diversifying the image regions used for indexing within the provenance image filtering task.

\item \emph{Iterative Filtering}: a novel querying strategy that iteratively retrieves images that are  directly or indirectly related to the query, considering all possible hosts, donors, composites, and their respective near duplicates.

\item \emph{Clustered Provenance Graph Construction}: a novel graph construction algorithm that clusters images according to their content (joining near duplicates into the same clusters), prior to establishing their intra- and inter-cluster relationship maps.

\item State-of-the-art results on the provenance analysis benchmark released by the American National Institute of Standards and Technology (NIST)~\cite{nist2017dataset}.

\item A new dataset of real-world scenarios containing composite images from Photoshop battles held on the Reddit website~\cite{reddit2017photoshopbattles}.
Experiments performed over this dataset highlight the real-world applicability of the approach.
\end{enumerate}

% Cut for space. WJS
%In the rest of this article, we explain what a practical end-to-end pipeline for provenance analysis looks like.
%In the rest of this article, we explain how the proposed %end-to-end pipeline for provenance analysis looks like.
%In Sec.~\ref{sec:rw}, we review the relevant literature %in this emerging area of image processing, highlighting %the advances made by existing approaches, as well as %their current weaknesses.
%Then, in Sec.~\ref{sec:prop}, we introduce methods for %provenance image filtering and graph construction.
%Aiming at assessing the performance of the proposed %solutions, a experimental setup is described in %Sec.~\ref{sec:expsetup}, while its results are reported %in Sec.~\ref{sec:results}.
%Finally, we provide some conclusions and discuss future %work in Sec.~\ref{sec:conc}.
 	            % Introduction
\section{Related Work}
\label{sec:rw}

%% 1. Provenance image filtering and CBIR
%In real-world scenarios, before finding associations %among elements of a relevant pool of images, there is the %necessity of building such pool by selecting samples from %a large corpus (\textit{e.g.}, the Internet), as long as %they are potentially related to a given query of %interest.
%Such activity constitutes the task of provenance image %filtering, which is related to the literature of %content-based image retrieval (CBIR).

% CBIR state of the art
\textit{Content-based image retrieval (CBIR).} 
In recent years, research advances in the domain of CBIR have included optimizing the memory footprint of indexing techniques and employing graphical processing units (GPU) for parallel search.
A recent technique proposed by Johnson~et~al.~\cite{johnson2017billion} utilizes state-of-the-art image indexing (Optimized Product Quantization (OPQ)~\cite{ge2013optimized}) and runtime optimization to perform similarity search on the order of a billion images.
Such approaches can be directly applied to perform image filtering for provenance analysis. 
However, as they follow the traditional CBIR inverted-file index pipeline~\cite{Zisserman_2003}, they will not generalize to all cases due to the nature of the problem.
While regular CBIR will probably retrieve good host candidates to the query, in the face of compositions (which are fairly common in provenance analysis), small donors will not be highly ranked (or will not even be retrieved) without adaptations to the base approach.

% Allan's ICIP paper
%In such a direction, Pinto~et~al.~\cite{pinto2017filtering} were the first to improve the retrieval of alien donors related to a query in the scope of multimedia phylogeny.
The work of Pinto et al.~\cite{pinto2017filtering} improves the retrieval of donors related to a query in the scope of provenance analysis.
The paper introduces a two-tiered search approach. The first tier constitutes a typical CBIR pipeline, while the second tier provides a context-aware query-masking technique, which selects the regions from the query that make it divergent from hosts previously obtained in the first tier.
With such regions as evidence, a second search is performed, this time avoiding hosts and retrieving additional potential donor images.
%Although the approach of Pinto~et~al. does improve the retrieval of alien donors, it adopts a very query-centric point of view, with respect to the problem of provenance analysis.
Although such an approach does improve the retrieval of donors, it adopts a very ``query-centric'' point of view with respect to the problem of provenance analysis.
It only finds the hosts and donors that directly share content with the query, ignoring the other descendants and the ancestors of such hosts and donors, which are indirectly related to the query.

% This paper's contribution upon the ICIP paper. Move this %to the letter?
%Nevertheless, a proper provenance analysis is expected to %retrieve the images that also potentially share content %with hosts and donors (\textit{i.e.}, their respective %hosts and donors), and the further hosts and donors, so %forth.
%For that reason, we extend their work by introducing the %technique of \emph{iterative filtering}, that aims at %solving such limitation.
%In addition, we also propose a novel interest point %detection heuristic that increases the probability of %describing --- and therefore indexing and searching --- %small alien regions.
%In summary, it spreads the detected interest points over %the image, therefore being called \emph{distributed %interest point selection}.
%Both techniques are detailed in %Sec.~\ref{sec:prop:filtering}.

%% 2. Provenance graph construction
% Common point: find associations among images
\textit{Image processing for image associations.}
In our proposed workflow, the filtering step yields relevant images, and then provenance graph construction is performed.
The provenance graph construction step involves finding diverse types of associations among images based on their similarities and/or dissimilarities.
For that reason, it is related to tasks such as visual object recognition~\cite{russakovsky_2015}, scene recognition~\cite{zhou_2017}, place recognition~\cite{lowry_2016}, object tracking~\cite{cehovin_2016}, near-duplicate detection~\cite{winkler_2013}, and image phylogeny~\cite{Dias_2012}, since they all rely on the comparison of two or more images.

% Generalization vs. Specialization tasks
Some visual association tasks may be general, as they relate images based on the common characteristics that optimally make them related.
This is the case, for instance, for object recognition. For example, a query image that implicitly requests ``retrieve all the images containing dogs'' may also be assumed to be generalized (any breed, color, or size). Scene recognition (\textit{e.g.}, ``retrieve all the images depicting bedrooms'') may also include generalized queries.
In such situations, a high content diversity among the related images is usually desired~\cite{Deselaers_2009}.
By contrast, some image association tasks may be specialized, in the sense that they aim at extracting the specific characteristics that aid in the visual identification of a sample in a particular setting.
That is the case of place recognition (\textit{e.g.}, retrieve all the images of Times Square), and object tracking (\textit{e.g.}, segment the target vehicle plate across the frames of a street surveillance video).
%In these cases, diversity is only acceptable regarding %different poses, views, and ages of the same object.

Techniques for associating images in a general way include comparing global image representations~\cite{Deselaers_2010, oliva_2006}, employing bags of visual features~\cite{avila_2013, nanni2013heterogeneous} and using convolutional neural networks (CNNs)~\cite{szegedy_2015, krizhevsky_2017, zhou2014learning, wang2017knowledge}.
Techniques for associating images in a specialized way include assessing local feature matching~\cite{maresca_2013, yang2015ransac, joly_2007, huang_2008,silva_2015}, image patch matching~\cite{milford_2014}, and evaluating the quality of image registration, color matching, and mutual information~\cite{Costa_2017, bharati2017uphy}.
Particularly, provenance analysis is by definition closer to the specialized tasks; for that reason, in this work, we benefit more from techniques mentioned in the latter group.

Although one can adapt deep CNNs to provenance analysis by optimizing them for specialization rather than generalization at training time, such a procedure is --- at the present time --- only accomplishable at the expense of prohibitive training times, the need for a reasonably large cluster of GPUs for model screening via hyperparameter optimization, and a sufficiently large amount of available training data~\cite{chollet_2017}.
In addition, making such a solution perform at scale at inference time is also challenging.
After running benchmark experiments using CNN-based approaches for finding image associations and noting long run-times, we have intentionally chosen to pursue faster alternatives to deep learning in this work.

%Particularly, provenance analysis is by definition closer to the latter group of specialization tasks.
%Although one can arguably adapt trendy deep CNNs to provenance analysis, by making networks present specialization capability rather than their typical generalization power, this would have to be done at currently prohibitive training costs, considering the needed processing power and large amount of available training data.
%In addition, making such a solution work at scale would also be challenging.
%For those reasons, in the present work, we benefit more from the techniques belonging to the latter group.

%\begin{table*}[t]
%\caption{TODO}
%\centering
%\begin{tabular}{ccccccccc}
%\hline
%Reference & Near duplicates? & Semantically similar? & Composites? & Assumed transformations & Pixel-wise comparison & Graph algorithm & Input & Output\\
%\hline
%\cite{Dias_2012} & Yes & No & No & cropping, warping, color correction, JPEG compression & MSE & oriented Kruskal & set of near duplicates & phylogeny tree\\
%\hline
%\end{tabular}
%\end{table*}

% Image phylogeny
\textit{Image phylogeny trees.}
Provenance analysis is related to the simpler task of image phylogeny, which seeks to recover a tree of relationships.
Kennedy and Chang~\cite{Kennedy_2008} were the first to point out the possibility of relying on the color information of pixels and local features for gathering clues about plausible parent-child relationships among images.
Based upon the pixel colors and local features, they suggest detecting a closed set of directed manipulations between pairs of content-related images (namely copy, scaling, color change, cropping, content insertion, and overlay detection).

%Rather than exhaustively modeling all the possible manipulations between near-duplicate images, Dias et al.~\cite{Dias_2011} departed from the assumption of having a good dissimilarity function that can be used for building a pairwise image dissimilarity matrix $D$.
Rather than exhaustively modeling all of the possible manipulations between near-duplicate images, Dias et al.~\cite{Dias_2011} suggest having a good dissimilarity function that can be used for building a pairwise image dissimilarity matrix $D$.
Accordingly, they introduce \emph{oriented Kruskal}, an algorithm that processes $D$ to output an \emph{image phylogeny tree}, a data structure that expresses the probable evolution of the near duplicates at hand.
In subsequent work, Dias et al.~\cite{Dias_2012} formally present the dissimilarity-calculation protocol that is widely used in the related literature for computing $D$.
They then go on to conduct a large set of experiments with this methodology, considering a family of six possible transformations, namely scaling, cropping, affine warping, brightness, contrast, and lossy content compression~\cite{Dias_2013_large}.
Finally, in~\cite{Dias_2013}, Dias et al. replace oriented Kruskal with other phylogeny tree building methods: best Prim, oriented Prim, and %Chu-Liu, Bock and
Edmonds' optimum branching~\cite{edmonds_1967}, with the last solution consistently yielding improved results.

\begin{figure*}[!htbp]
\centering
\includegraphics[width=18.5cm]{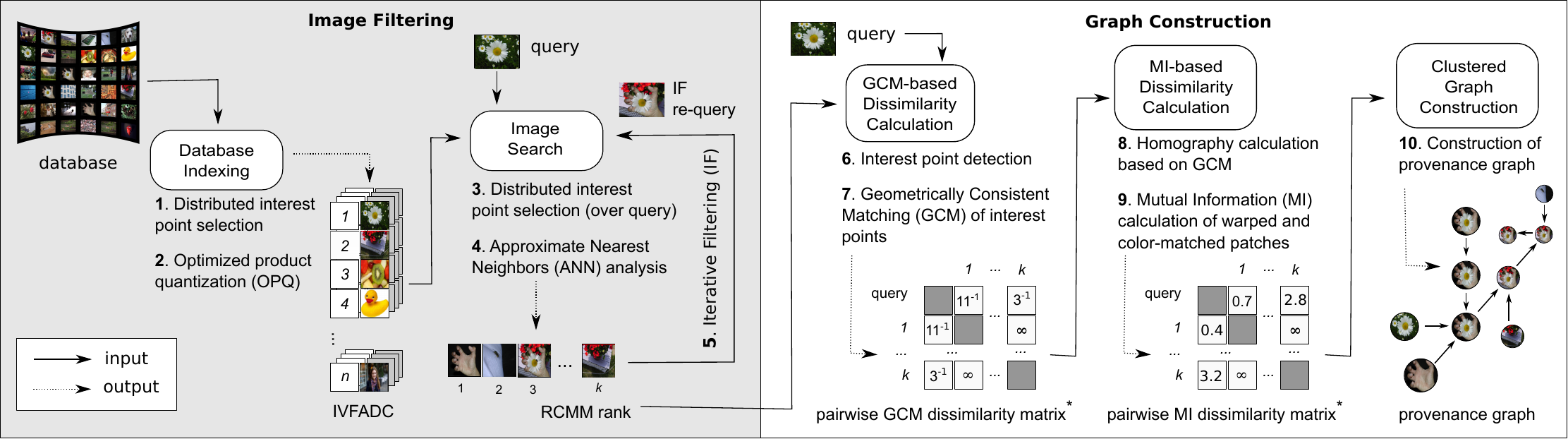}
\caption{Proposed pipeline for end-to-end provenance analysis.
The sequence of activities is divided into two parts, which address the tasks of \emph{image filtering} (left panel) and of \emph{graph construction} (right panel).
IVFADC stands for \emph{Inverted File System with Asymmetric Distance Computation}.
RCMM stands for \emph{Reciprocal Condition Matching Measure}.
The value of $n$, within IVFADC, is the number of images in the database.
The value of $k$ is a parameter of the solution and is related to the size of the RCMM rank used for building the provenance graph.
$^\ast$Reported values are merely illustrative.
}
\label{fig:solution}
\end{figure*}

% Image phylogeny: presence of semantically similar images
\textit{Image phylogeny forests.}
The image phylogeny solutions mentioned up to this point were conceived to handle near duplicates; they do not work in the presence of semantically similar images.
Aware of such limitations, Dias et al.~\cite{Dias_2013_fsi} extend the oriented Kruskal solution to \emph{automatic oriented Kruskal}, an algorithm that finds a family of disjoint phylogeny trees (a phylogeny forest) from a given set of near duplicates and semantically similar images, such that each tree describes the relationships of a particular group of near duplicates.
Analogously, Costa et al.~\cite{Costa_2014} provide two extensions to the optimum branching algorithm, namely \emph{automatic optimum branching} and \emph{extended automatic optimum branching}, both based on automatically calculated cut-off points.
%The former solution, similarly to automatic oriented Kruskal, employs a dissimilarity-threshold-based scheme to decide whether to add a candidate image to the tree currently being built, or to leave it to the next one.
%The latter solution, in turn, automatically finds the disjoint phylogeny trees, at the cost of performing more computations.
Alternatively, Oikawa et al.~\cite{oikawa_2016} propose the use of clustering techniques for finding the various phylogeny trees; the idea is to group images coming from the same source, while placing semantically similar images in different clusters.
Finally, Costa et al.~\cite{Costa_2017} improve the creation of the dissimilarity matrices, regardless of the graph algorithm used for constructing the trees.

% Image phylogeny: multiple parents
\textit{Multiple parenting phylogeny trees.}
Although previous phylogeny work established preliminary analysis strategies and algorithms to understand the evolution of images, the key scenario of image composition, in which objects from one image are spliced into another, was not addressed.
Compositions were first addressed within the phylogeny context by Oliveira et al.~\cite{Oliveira_2016}.
The solution presented by these authors assumes two parents (one host and one donor) per composite.
Extended automatic optimum branching is thus applied for the construction of ideally three phylogeny trees: one for the near duplicates of the host, one for the near duplicates of the donor, and one for the near duplicates of the composite.
%Moreover, they adopt the same dissimilarity-calculation protocol proposed in~\cite{Dias_2012}~and~\cite{Dias_2013}, with the fixed set of possible transformations.
Even though this work is very relevant to ours herein, it has a couple of limitations.
First, it does not consider the possibility of more than two images donating content towards one composite image (such as the composite with sharks in Panel B of Fig.~\ref{fig:teaser}).
Second, Oliveira et al. require all images to be in JPEG format.

% From image phylogeny to provenance analysis
\textit{Provenance graphs.}
To date, the entire image phylogeny literature has made use of metrics that focus on finding the root of the tree, rather than evaluating the phylogeny tree as a whole, considering every image transformation path in the case of provenance.
Aware of such limitations and aiming to foster more research on the topic, NIST has recently introduced new terminology, metrics, and datasets, coining the term \emph{image provenance} to express a broader notion of image phylogeny, and suggesting directed acyclic \emph{provenance graphs}, instead of trees, as the data structure that describes the provenance of images~\cite{nist2017plan}.
They also suggest the use of a \emph{query} as the starting point for provenance analysis.

%Following this, Bharati et al.~\cite{bharati2017uphy} introduced a more generalized method of provenance graph construction, which does not assume anything about the images and content transformations.
Following this, Bharati et al.~\cite{bharati2017uphy} introduced a more generalized method of provenance graph construction, which does not assume anything about the images and transformations.
A content-based method for the construction of undirected provenance graphs is proposed, which relies upon the extraction and geometrically-consistent matching of interest points.
Utilizing this information to build the dissimilarity matrix, the method uses Kruskal's %spanning tree
algorithm to obtain the provenance graph.
The approach performs well over small cases, even in the presence of distractors (\textit{i.e.}, images that are not related to the query). %, but the performance degrades for larger cases.
%The paper concludes that the method holds potential to generalize well for real-world provenance cases.

% Too much defense?
%This work directly benefits from the methods presented %in~\cite{bharati2017uphy}.
%However, differently from that one, it provides a %complete end-to-end pipeline for image provenance %analysis, including the initial filtering step.
%In addition, it introduces a novel graph algorithm for %constructing provenance graphs (clustered provenance %graph expansion), a strategy for obtaining directed %edges, and results over more datasets.
%Table~\cite{table:ref} summarizes the literature of image %phylogeny and provenance analysis, with this work put in %perspective.

%% Removing this to save space (this discussion isn't 
%% immediately relevant to our paper) WJS
%% Types of image associations and our focus
%\textit{Beyond visual-content associations.}
%This work and the aforementioned literature focus on %finding spatial associations among images, relying solely %upon the visual content (\textit{e.g.}, matched interest %points and regions).
%There is some prior work aimed at comparing images based %upon information other than pixels.
%That is the case, for instance, of the establishment of %temporal associations for photo %sequencing~\cite{basha2012photo}, and the analysis of %image meta-data for manipulation %detection~\cite{fan_2011}.
                    % Related Work
\section{Provenance Analysis Methodology}
\label{sec:prop}

% Preamble
As described in Sec.~\ref{sec:intro}, the task of image provenance analysis is divided into two major steps, namely \emph{Provenance Image Filtering} and \emph{Provenance Graph Construction}.
Fig.~\ref{fig:solution} depicts an overview of the proposed solution in this context.

\subsection{Provenance Image Filtering}
\label{sec:prop:filtering}
The problem of image filtering for the provenance task is different from the typical image retrieval task: a given query image may %simultaneously
fulfill one or both of the following conditions:

\begin{itemize}
    \item The query may have a relationship to various \emph{near duplicates}. The near duplicates may be \emph{hosts} of the query (in the case of the query being a composite that inherits the background from a near duplicate) or the query itself may be a host, as in the case of the query donating a background to the near duplicates.
    
    \item The query may be a composite with a relationship to one or more donors, whose content may be entirely disjoint.
    Donors can even be composites themselves, with their own hosts and donors.
\end{itemize}

In such scenarios, the retrieval method must return as many of the directly and indirectly related images as possible.
These aspects define a unique image retrieval and filtering problem, known as \emph{Provenance Image Filtering}~\cite{pinto2017filtering, nist2017plan}, which is different from more typical \emph{near-duplicate} or \emph{semantically similar} image retrieval.
%Therefore, a new system must be designed with the specific purpose of solving such a task.
In this work, we assume that a ground-up system must be deployed for search, retrieval, and filtering, instead of relying on currently available resources such as \emph{Google}~\cite{barroso2003web} or \emph{TinEye}~\cite{jacquicheng}.
%As depicted in Fig.~\ref{fig:solution}, the proposed system starts with a modified interest point selection method (detailed in %Sec.~\ref{sec:prop_spread_kp}), whose output is fed to %a highly-parallelizable
%an image index computation approach inspired by the work in~\cite{johnson2017billion} (detailed in Sec.~\ref{sec:prop_indxing}).
%Once the image indices are obtained, the image search is performed in a way close to~\cite{johnson2017billion} (which is explained in %Sec.~\ref{sec:prop_search}), with the additional step of \emph{iterative filtering} (explained in %Sec.~\ref{sec:prop_iterative_filtering}) that adapts image filtering to the provenance scenario.
%Finally, in Sec.~\ref{sec:prop_scaling}, we explain the strategy to handle millions of images.
 
\vspace{0.2cm} 
\subsubsection{Distributed Interest Point Selection} 
\label{sec:prop_spread_kp}
Due to the nature of the manipulations seen in tampered images, it is important to build a filtering system that is tolerant to a wide range of image transformations.
Hence, we adopt a low-level image representation that is based on interest points and local features, since they are reportedly tolerant to transformations such as scaling, rotation, and contrast adjustment~\cite{Bay:CVIU:2008}.
Nevertheless, while regular interest points are mostly designed to identify corners and blobs on the image, we also want to describe and further index homogeneous areas with low response and consequently a sparse amount of detected interest points, for retrieving images with the same type of content.
Although one can use a dense sampling approach to extract interest points within those regions, this is computationally prohibitive in the context of searching millions of images~\cite{pinto2017filtering}.

Therefore, we introduce a new method called \emph{distributed interest point selection} that aims at keeping a sparse approach while being able to provide interest points inside low-response areas.
For that, we extend Hessian-based detectors (such as Speeded-Up Robust Features (SURF)~\cite{Bay:CVIU:2008}) in the following way.
Instead of employing a threshold $t$ to collect interest points whose local Hessian values are greater than $t$, we define a parameter $p$ that expresses the fixed amount of interest points we want to extract from each target image.
Within these $p$ interest points, $m < p$ interest points are extracted for the reason of being the top-$m$ regions with the $m$ strongest Hessian values.
The remaining $n = p - m$ are extracted from the set containing the post-top-$m$ interest points, which is also sorted according to the Hessian response.
Starting from the $(m+1)$-th strongest interest point, we only add the current interest point if it does not \emph{overlap} with another already selected interest point; otherwise, we try to add the next strongest interest point, up to the point of obtaining $n$ interest points.

\begin{figure}[!t]
\centerline{
    \subfloat[]{\frame{\includegraphics[width=4cm]{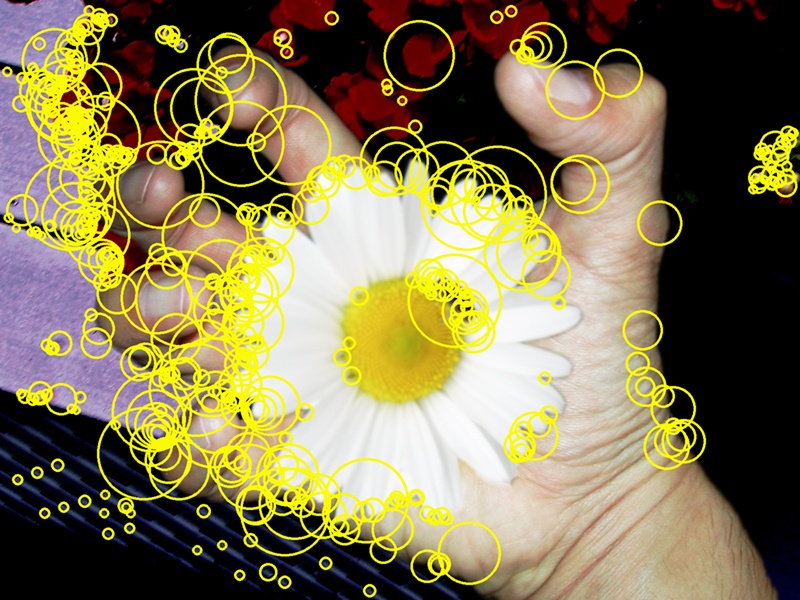}}}
    \hfil
    \subfloat[]{\frame{\includegraphics[width=4cm]{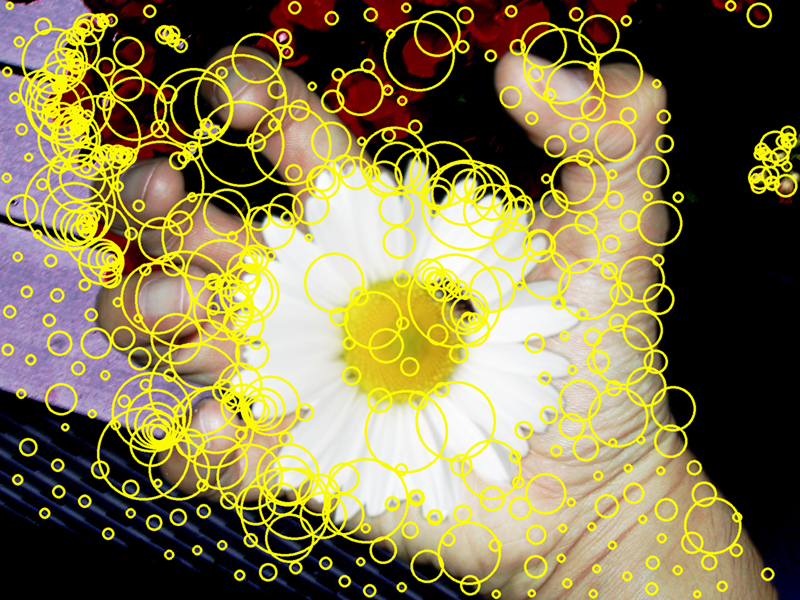}}}
}
\caption{Effects of using the approach of distributed interest point selection.
In (a), the result of a regular SURF interest point detection.
In (b), the result of the distributed approach over the same image, with many more points over homogeneous regions, such as the skin of the wrist.}
\label{fig:disitributed_kp}
\end{figure}

Fig.~\ref{fig:disitributed_kp} depicts the effect of using the distributed approach along with SURF.
Fig.~\ref{fig:disitributed_kp}~(a) depicts a regular SURF detection, while Fig.~\ref{fig:disitributed_kp}~(b) depicts the distributed version, over the same image.
Fig.~\ref{fig:disitributed_kp}~(b) presents more points over the skin of the wrist and background (which are more homogeneous regions) than Fig.~\ref{fig:disitributed_kp}~(a).

%Taking into consideration that each interest point is actually a circular blob with a center $(x, y)$ and a radius $r$, we establish that two points overlap if they do not collide (\textit{i.e.}, they do not share describable content).

\vspace{0.2cm} 
\subsubsection{Database Indexing}
\label{sec:prop_indxing}
The next step is to build the image index structure. %, which we refer to as the CBIR system.
After interest point detection and feature extraction, we are left with $p$ description vectors per image.
For an image collection $C$:

\begin{equation}
C_i,\quad s.t.\quad  i \in \mathcal{I}_{img} = \{0, 1, \ldots, |C|\},
\end{equation}

\noindent our subsequent feature collection is:

\begin{equation}
F_i \quad s.t.\quad  i \in \mathcal{I}_{ind} = \{0, 1, \ldots, |C| \times p\},
\end{equation}

\noindent where $\mathcal{I}_{img}$ denotes the numbered index set of full images within $C$, and $\mathcal{I}_{ind}$ indicates the subsequent numbered index set assigned to individual features in $F$.
We transform $F$ to a new space using \emph{Optimized Product Quantization} (OPQ)~\cite{ge2013optimized} to make the feature space well-posed for coarse \emph{Product Quantization} (PQ).
We refer to this new rotated feature set as $F_{r}$.
From a random sample of $F_{r}$, a coarse set of representative centroids $S$ is generated using PQ.
A subsequent \emph{Inverted File System with Asymmetric Distance Computation} (IVFADC)~\cite{jegou2011product}) is generated from $S$, allowing for fast and efficient search.

\vspace{0.2cm} 
\subsubsection{Image Search} % and Retrieval}
\label{sec:prop_search}
%Once the CBIR system is built, it can be searched via feature-wise queries.
Once the database images are indexed, a search procedure can be performed via feature-wise queries.
For a query image $Q$, a set of $p$ distributed SURF features $F_{d}$ is extracted and submitted to the system.
Each image $Q$ returns a matrix of indices of \emph{Approximate Nearest Neighbors} (ANN) $R$ of size $(p \times K)$.
The $R_{ij}$ value is computed using \emph{Asymmetric Distance Computation} (ADC)~\cite{jegou2011product}, where $i$ denotes the $i$-th query feature of $Q$, and $j$ denotes the $j$-th ANN index of the $i$-th query feature within $F_{d}$:

\begin{equation}
\begin{aligned}
& R_{i,j}=\zeta(F_{d_{i}})_{j} \in \mathcal{I}_{ind},\ s.t.\\
& i \in \{0,1,\ldots,|F_{d_{i}}|\}\ \text{and}\  j \in K,
\end{aligned}
\end{equation}

\noindent where $\zeta$ signifies a single query on the %CBIR
filtering system, and $K$ is the parameter of the K-nearest neighbors for %the CBIR
the system to return.
Once the set $R$ is calculated, we map $R$ from the $\mathcal{I}_{ind}$ space to the $\mathcal{I}_{img}$ space.
The number of unique image indices is computed as:

\begin{equation}
R_{img}=\xi(R,\mathcal{I}_{ind},\mathcal{I}_{img}).
\end{equation}
% TODO: Daniel: this last step is not clear to me.

Once $R_{img}$ is obtained, a sorted set of votes is calculated for representing the final global query results of $Q$:

\begin{equation}
\begin{aligned}
& V_{i+1}=argmax_{x}\{\phi(x,R_{img})-V_{i}\},\ s.t. \\
& x \in \theta(R_{img})=\{R_{img}\}\ \text{and}\  V_0 = 0.
\end{aligned}
\end{equation}

The function $\xi(R,\mathcal{I}_1,\mathcal{I}_2)$ maps index values in $R$ to the $\mathcal{I}_{img}$ index domain, allowing each $R_{ij}$ to represent the image it belongs to. 
The $\phi(x,R)$ value is an accumulator that returns the tally of all values of $x$ within $R$.
The $\theta(R)$ value represents the set of distinct values within $R$.

Using this scheme, we are able to retrieve images that only partially match $Q$, even in the presence of many noisy matches. Small objects will have high chances of accumulating values while spurious interest points will not.

\vspace{0.2cm} 
\subsubsection{Iterative Filtering}
\label{sec:prop_iterative_filtering}
Once a first rank of images is retrieved through the search algorithm, we iteratively refine the results to add images that are not directly related to the query, but are still related in some way to its provenance.
%Refer to Fig.~\ref{fig:solution} for an illustration of this process.

In contrast to the approach described by Pinto et al.~\cite{pinto2017filtering}, which employs a two-tiered search to retrieve the small donors of the query after masking the regions that diverge between the query and the first images of the retrieved rank, in this work we employ the \emph{reciprocal condition matching measure} (RCMM) proposed in~\cite{brogan2017spotting} to identify and suppress the near duplicates of the query.
%Given that, the greater the RCMM between two images, the greater the possibility of them being near duplicates, we suppress the retrieved images whose RCMM values with the query are high.
Given that a large RCMM value between two arbitrary images indicates that they are probably near duplicates, we suppress the retrieved images whose RCMM values with the query are large.
% TODO: Daniel: ask Joel more details about this suppression. Is it based on a threshold? Maybe rank position?
The non-suppressed (and therefore non-near-duplicate) images of the current rank are then provided as new queries to the next search iteration, which is performed using the same method explained in Sec.~\ref{sec:prop_search}.

%By applying the above process for a number of iterations, we search various sets of non-near-duplicate queries (which are potentially donors) and end up with a set of ranks, which are then flattened and re-ranked using RCMM.
%In the end, we obtain a less query-centric rank of images, which contains the images directly and indirectly related to the query.
%As we show through experiments in Sec.~\ref{sec:results}, such a strategy improves the recall of the provenance image filtering task, when compared to the application of a regular CBIR system.

By applying the above process for a number of iterations, we search various sets of non-near-duplicate queries (which are potentially donors) and end up with a set of ranks, which are then flattened and re-ranked using RCMM.
In the end, we obtain a less query-centric rank of images, which contains not only images directly related to the query, but also indirectly related (\textit{e.g.}, ancestors of the donors of the query).
As will be demonstrated in Sec.~\ref{sec:results}, such a strategy improves the recall of the provenance image filtering task.

% TODO: Daniel: Ask Joel to review the entire following section, including images.
\vspace{0.2cm} 
\subsubsection{Large-Scale Infrastructure}
\label{sec:prop_scaling}

\begin{figure}[t]
\centering
\includegraphics[width=9cm]{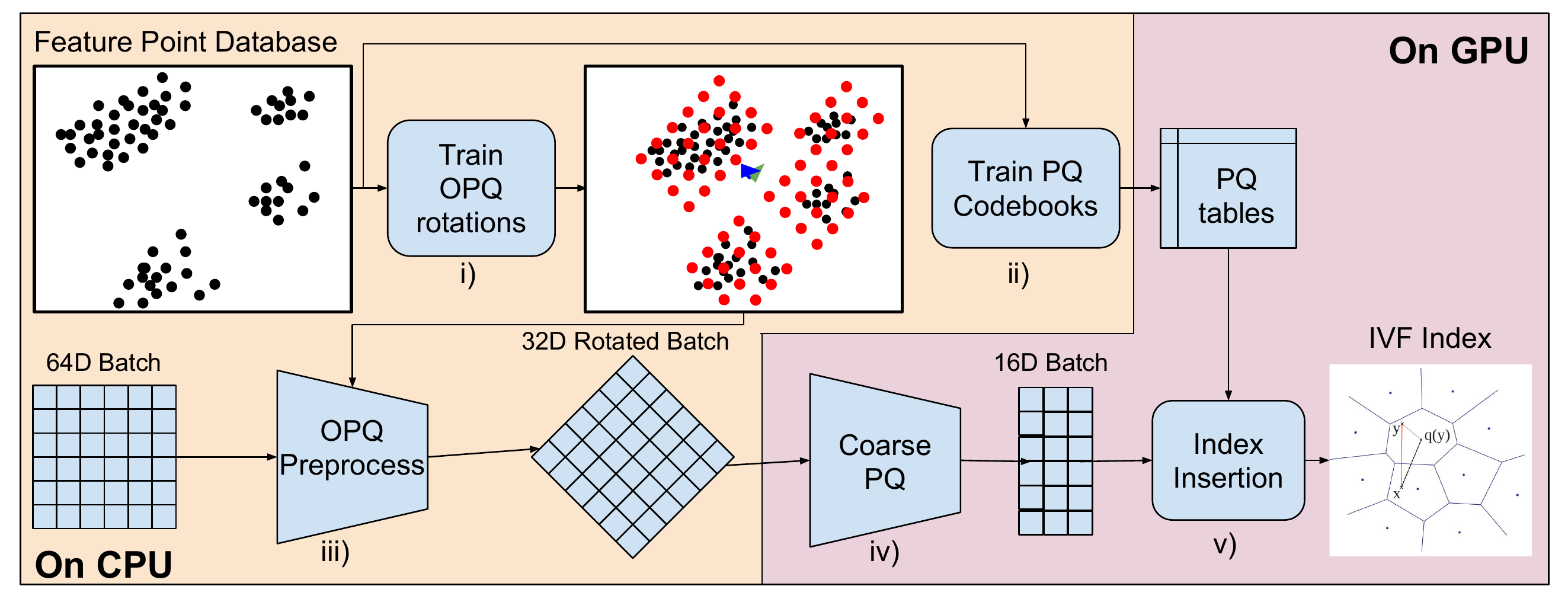}
\caption{Filtering pipeline infrastructure.
The orange area (left) shows computations that are performed on a CPU.
The purple area (right) shows the index ingestion steps that are performed on a GPU.}
\label{fig:filterPipeline}
\end{figure}

Fig.~\ref{fig:filterPipeline} shows the proposed full pipeline for index \emph{training} and \emph{construction} (previously explained in Sec.~\ref{sec:prop_indxing}).
Index training refers to the process of learning the OPQ rotations and PQ codebooks from a sampling of the local features that are extracted from the target dataset.
%We propose to perform such a step ahead of time with typical central processing units (CPU).
Index construction, in turn, refers to the computation of the inverted file indices, after properly rotating the previously extracted local features.
The learning of OPQ rotations and PQ codebooks can be done in advance on a CPU, but the construction of indices is well suited to the capabilities of graphical processing units (GPU), allowing for faster computation.
% TODO: Prof. Rocha's note: Clarify this last sentence. Make it clear what you did in CPU and in GPU.  

% TODO: Prof. Scheirer's note: make on CPU and on GPU labels bold.

\begin{figure}[!htbp]
\centering
\includegraphics[width=9cm]{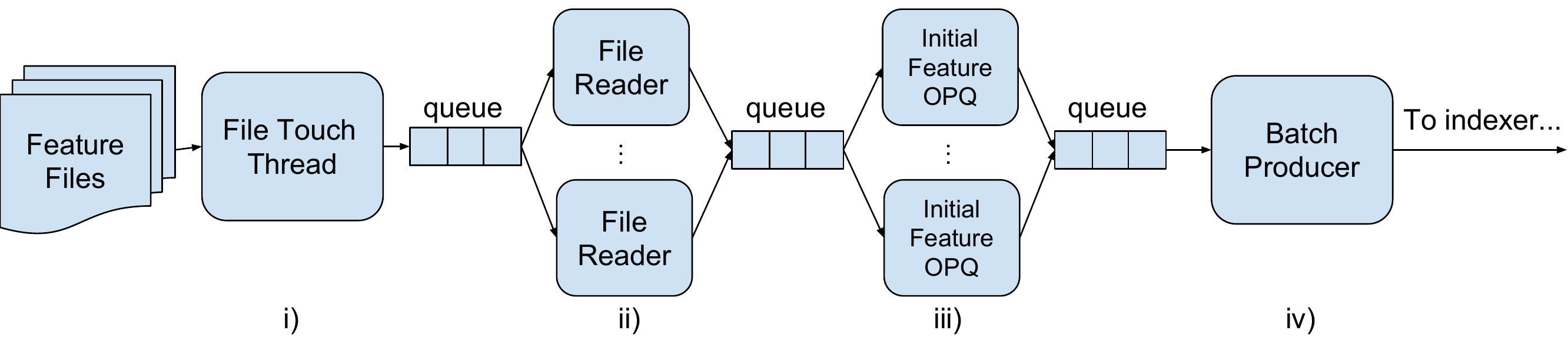}
\caption{
Producer-consumer index ingestion.
Each file contains features for an image.
These file locations are pre-loaded into cache via a rate-limited ``touch" thread, and are read on a producer-consumer multi-threaded basis.}
\label{fig:touchPipeline}
\end{figure}
% TODO: Prof. Scheirer's note: make image texts larger.

Besides employing GPUs to efficiently build and search an index of over 1 million high-resolution images, additional steps must be taken to increase the pipeline speed.
To date, most indexing algorithms require singular large files containing all features to be ingested at once \cite{flann_pami_2014,Attach:binary_matching_crv2012}, either due to implementation choices or algorithm limitations.
The operation of concatenating all features from a set of images into a single file is prohibitively time consuming when dealing with more than a few million interest points.
Because our scenarios require the ingestion of multiple billions of interest points, a different solution must be adopted, in order to avoid the need for file concatenation.
For that, we propose a multi-threaded producer-consumer setup, as shown in Fig.~\ref{fig:touchPipeline}.
In our pipeline, we provide a single feature file per image.
The pipeline begins with the ``touch" thread, which systematically loads image feature file locations into the computer's file system cache, for faster retrieval in later stages.
Then, a reading thread takes touched files and loads them into memory.
A third thread takes sets of loaded feature files and produces feature batches of size $B$ that are optimized in size for GPU ingestion.
The fourth thread applies the initial OPQ pre-processing rotations to the feature set, before sending the final batch to the GPU. Using this method, we are able to process billions of features from high-resolution image datasets orders of magnitudes faster than previous methods.

\subsection{Provenance Graph Construction}
\label{sec:prop:graph_building}

As one can observe in Fig.~\ref{fig:solution}, the provenance graph construction task builds upon the image rank that is obtained by the provenance image filtering task, and ends up with the provenance graph.
Therefore, at this point, we can assume that (in the best scenario) all images directly and indirectly related to the query are available for constructing the provenance graph, as well as some \emph{distractors} (images that should not be present in the provenance graph, because they are not related to any of the images within it).

The presence of distractors at this step is more of a matter of design.
Taking into consideration that, in~\cite{bharati2017uphy}, experiments show distractors not impacting the provenance graph construction too much, and aiming to keep the provenance image filtering part as simple as possible, we give the subsequent dissimilarity matrix calculation task the duty of removing distractors. Therefore, the input is a set containing the $k$ top-retrieved images and the query, which are then used for building dissimilarity matrices.

\vspace{0.2cm}
\subsubsection{Calculation of Dissimilarity Matrices}
\label{sec:prop_dismat}
Similar to~\cite{Dias_2012}, given the set $I$ containing the $k$ top-retrieved images and the query, a dissimilarity matrix $D$ is a $(k+1)\times(k+1)$~matrix whose elements $d_{ij}$ describe the dissimilarity between images $I_i$ and $I_j$, respectively the $i$-th and $j$-th images of $I$.
Depending on how the values $d_{ij}$ are calculated, $D$ can be either symmetric or asymmetric.

In this work, following the solution proposed in~\cite{bharati2017uphy}, we neither make any strong assumptions with respect to the transformations that might have been used to generate the elements of $I$, nor impose limitations on the presence of near duplicates, semantically similar images, or multi-donor composites.
Instead, we focus on analyzing the shared visual content between every pair of images $(I_i, I_j)$ through two ways of calculating $d_{ij}$.
In the first one, we set $d_{ij}$ as the inverse of the number of geometrically-consistent interest-point matches (GCM) between images $I_i$ and $I_j$; in this particular case, the matrix $D$ is symmetric.
In the second one, we set $d_{ij}$ as the \emph{mutual information} (MI) between a color transformation $T_j(I_i)$ of image $I_i$ towards image $I_j$; in this case, the matrix $D$ is asymmetric.
Both methods are described below.

\vspace{0.2cm}
\subsubsection*{GCM-based dissimilarity}
%As illustrated in Fig.~\ref{fig:solution},
Provenance graph construction starts with the detection of interest points over each one of the $k+1$ images that belong to $I$.
At this step, different interest point detectors can be applied, such as SURF~\cite{Bay:CVIU:2008} or Maximally Stable Extremal Regions (MSER)~\cite{Matas_2004}, with each one yielding a particular dissimilarity matrix. Once the interest points are available and properly described through feature vectors (\textit{e.g.}, SURF features~\cite{Bay:CVIU:2008}), we find correspondences among them for every pair of images $(I_i, I_j)$.
Let $P_i$ be the set of feature vectors obtained from the interest points of image $I_i$, and $P_j$ be the set of features obtained from $I_j$.
For each feature belonging to $P_i$, the two best matching features are found inside $P_j$ using Euclidean distance (the closer the features, the better the match).
Inspired by Nearest-Neighbor-Distance-Ratio (NNDR) matching quality~\cite{lowe2004distinctive}, we ignore all the features whose ratio of the distances to the first and to the second best matching features is smaller than a threshold $t$, since they might present a poor distinctive quality.
The remaining features are then kept and finally matched to their closest pair.

Even with the use of NNDR, it is not uncommon to gather \emph{geometrically inconsistent matches}, \textit{i.e.}, contradictory interest-point matches that, if together, cannot represent plausible content transformations of image $I_i$ towards image $I_j$, and vice-versa. %, in terms of translation, rotation, and scaling.
%For the sake of illustration, contradictory matches regard the ones that cross each other, as one might observe through the highlighted matches in Fig.~\ref{fig:solution}.
To get rid of these matches, we adopt a solution that is able to build a geometrically-consistent model of expected interest-point positions from any pair of matches between images $I_i$ and $I_j$.
For example, consider two arbitrary matches $m_1$ and $m_2$, which respectively connect points $p_1 \in I_i$ and $q_1 \in I_j$, and points $p_2 \in I_i$ and $q_2 \in I_j$.
Based upon the positions, the distance $l_p$, and the angle $\alpha_p$ between points $p_1$ and $p_2$ (both from image $I_i$), as well as upon the positions, the distance $l_q$, and angle $\alpha_q$ between points $q_1$ and $q_2$ (both from image $I_j$), we estimate the scale, translation, and rotation matrices that make $p_1$ and $p_2$ respectively coincide with $q_1$ and $q_2$.
With these matrices, we transform every matched interest point of $I_i$ onto the space of $I_j$.
As one might expect, points that do not coincide with their respective peers after the transformations have their matches removed from the set of geometrically consistent matches.

% Daniel: I removed the text below to give more space for a better explanation of mutual-information-based solutions.
%Depending on the selected pair of matches for estimating the transformation model, though, different sets of consistent matches may be obtained.
%In this case, one should take the model that returns the set of consistent matches with the largest cardinality (\textit{i.e.}, the maximal set of consistent matches).
%In order to avoid performing an exhaustive search for such a set (by evaluating all the possible pairs of matches between images $I_i$ and $I_j$) we adopt a near-maximal approach, which evaluates only a subset of all the available matches.
%Let $m$ be the total amount of matches established between images $I_i$ and $I_j$, after the application of NNDR.
%After sorting the $m$ matches according to their distances, from the closest to the farthest, we build the mentioned subset with only the $n\leq m$ closest matches.
%With this relaxation, we cannot guarantee that the obtained set of consistent matches is maximal, unless we adopt $n=m$.
%However, there is a big chance of finding the maximal set, since it probably contains at least one of the $n$ best matches.

Finally, we compute the dissimilarity matrix $D$  by setting every one of its $d_{ij}$ elements as the inverse of the number of found geometrically-consistent matches between images $I_i$ and $I_j$.
In this case, the dissimilarity matrix is symmetric. %, since $d_{ij}=d_{ji}$.

\vspace{0.2cm}
\subsubsection*{MI-based dissimilarity}
The mutual-information (MI)-based dissimilarity matrix is  an extension of the GCM-based alternative (see Fig.~\ref{fig:solution}).
After finding the geometrically consistent interest-point matches for each pair of images $(I_i, I_j)$, the obtained interest points are used for estimating the homography $H_{ij}$ that guides the registration of image $I_i$ onto image $I_j$, as well as the homography $H_{ji}$ that analogously guides the registration of image $I_j$ onto image $I_i$.

In the particular case of $H_{ij}$, for calculating $d_{ij}$, after obtaining the transformation $T_j(I_i)$ of image $I_i$ towards $I_j$, $T_j(I_i)$ and $I_j$ are properly registered, with $T_j(I_i)$ presenting the same size of $I_j$, and the matched interest points relying on the same position.
We thus compute the bounding boxes that enclose all the matched interest points, within each image, obtaining two correspondent patches $R_1$, within $T_j(I_i)$, and $R_2$, within $I_j$.
As in~\cite{bharati2017uphy}, the distribution of the pixel values of $R_1$ is matched to the distribution of $R_2$, prior to calculating the pixel-wise amount of residual between them with MI.

From the point of view of information theory, MI is the amount of information that one random variable contains about another.
From the point of view of probability theory, it measures the statistical dependence of two random variables.
In practical terms, assuming each random variable as respectively the aligned and color-corrected patches $R_1$ and $R_2$, the value of MI is given by the entropy of discrete random variables:

\begin{equation}
\begin{aligned}
& MI(R_1, R_2) = \\
& \sum_{x \in R_1}^{ }{\sum_{y \in R_2}^{ }{p(x, y)}} \log \left ( \frac{p(x, y)}{\sum_{x}^{ } {p(x, y)} \sum_{y}^{ } {p(x, y)}} \right ),
\end{aligned}
\end{equation}

\noindent where $x \in [0,\ldots,255]$ refers to the pixel values of $R_1$, and $y \in [0,\ldots,255]$ refers to the pixel values of $R_2$.
The $p(x, y)$ value regards the joint probability distribution function of $R_1$ and $R_2$.
As explained in~\cite{Costa_2017}, it can be approximated by:

\begin{equation}
p(x, y) = \frac{h(x, y)}{\sum_{x, y}^{ } {h(x, y)}},
\end{equation}

\noindent where $h(x, y)$ is the joint histogram that counts the number of occurrences for each possible value of the pair $(x, y)$, evaluated on the corresponding pixels for both patches $R_1$ and $R_2$.
As a consequence, MI is directly proportional to the similarity of the two patches.

Back to $H_{ji}$, it is calculated in an analogous way of $H_{ij}$.
However, instead of $T_j(I_i)$, $T_i(I_j)$ is manipulated for transforming $I_j$ towards $I_i$. Further, the size of the registered images, the format of the matched patches, and the matched color distributions are different, leading to a different value of MI for setting $d_{ji}$.
As a consequence, the resulting dissimilarity matrix $D$ is asymmetric, since $d_{ji}\ne d_{ji}$.

\vspace{0.2cm}
\subsubsection*{Avoiding distractors}
As we have mentioned before, the image rank given to the provenance graph construction step may contain distractors, which need to be removed during the dissimilarity matrix calculation step.
When computing the dissimilarity matrix $D$, the solution proposed by Bharati et al.~\cite{bharati2017uphy} establishes matches between every pair of available images, including distractors.
By interpreting $D$ as the adjacency matrix of a multi-graph whose nodes are the images, they identify distractors as the nodes weakly connected (\textit{i.e.,} that present a small number of matches, down to none) to the minimum spanning tree that contains the query.
Assuming $(k + 1)$ as the number of image nodes, they perform $(k^2 + k) / 2$ operations to populate $D$.

In this work, we improve that process by the means of an iterative approach, which starts from the node of the query and then computes the geometrically consistent matches with the remaining $k$ images.
A set with only the strongly connected nodes is thus saved for the next iteration.
In the following iterations, the algorithm keeps trying to establish matches starting from the last set of strongly matched images, up to the point where no more strong matches are found.

Although simple, this solution may provide a significant improvement in the runtime of the dimissimilarity matrix calculation.
Let $d\leq k$ be the amount of distractors inside the image rank.
We avoid $(d^2 - d) / 2$ operations by applying the iterative solution.
In the case of a rank with 50 images ($k=50$), for instance, and 40 distractors ($d=40$) (indicating that the provenance graph contains only ten images), the number of operations is reduced from $1,275$ to $795$, significantly speeding up the runtime in case of small graphs.

\vspace{0.2cm}
\subsubsection{Clustered Provenance Graph Construction}
\label{sec:prop_cluster}
Once the GCM- and MI-based dissimilarity matrices are available, we rely on both for constructing the final provenance graph, by the means of a novel algorithm, named \emph{clustered provenance graph expansion}.
The main idea behind such a solution is to group the available images in a way that only near duplicates of a common image are added to the same cluster.

Starting from the image query $I_q$, the remaining images are sorted according to the number of geometrically consistent matches shared with $I_q$, from the largest to the smallest.
The solution then clusters probable near duplicates around $I_q$, as long as they share \emph{enough content}, which is decided based upon the number of matches.
After automatically adding the first image of the sorted set to the cluster of $I_q$, the solution iteratively analyzes the remaining available images.
For deciding if the $i$-th candidate image $I_i$ (where $i>1$) is a near duplicate, the algorithm keeps track of the number of matches $m_i$ between $I_i$ and the last image $I_{i-1}$ added to the cluster.
Let $\mu_{i-1}$ be the average number of matches of the cluster, and $\sigma_{i-1}$ be the standard deviation.
$I_i$ is connected to $I_{i-1}$ in the final provenance graph, if $m_1 \in [\mu_{i-1} \text{ - } \sigma_{i-1}; \mu_{i-1} \text{ + } \sigma_{i-1}]$; in such a case, $I_i$ is added to the cluster by affinity, and novel values of $\mu_i$ and $\sigma_i$ are calculated, for evaluating the next candidate $I_{i+1}$.
Otherwise, the current cluster is considered finished up to $I_{i-1}$.

As a consequence, the obtained clusters have their images sequentially connected into a single path, without branches.
That makes sense in scenarios involving sequential image edits where one near duplicate is obtained on top of the other, as in~\cite{nist2017dataset}.
To determine the direction of the entire path, we assume the dominant direction within all the edges that make part of the path.
To determine the direction of a single edge, we rely on the mutual information.
Let $D$ be the MI-based dissimilarity matrix, and consider two images $I_i$ and $I_j$, whose respective $D$ elements are $d_{ij}$ and $d_{ji}$.
As explained in~\cite{Oliveira_2016}, an observation of $d_{ij} > d_{ji}$ means that $I_i$ probably generated $I_j$.

Finally, whenever a cluster is finished and there are still disconnected available images, we find the image already added to the provenance graph whose number of matches with the remaining ones is largest.
This image is then assumed as the new query $I_q'$, over which the aforementioned clustering algorithm is executed, considering only the yet disconnected images.
As a result, the final provenance graph sees a branch rising from $I_q'$ as an orthogonal path containing new images.

% Conclusion of section
%In the following sections, we evaluate the proposed %provenance analysis pipeline, by detailing the %experimental setup (in Sec.~\ref{sec:expsetup}) and %reporting the experimental results over three interesting %datasets (in Sec.~\ref{sec:results}).                  % Proposed Solution
% Daniel's notes: this is a new version of the experimental setup, which was built upon Aparna and Joel's.
% Please refer to ../backup/04_expsetup_aparna_joel.tex for the original version.

\section{Experimental Setup}
\label{sec:expsetup}
Here we describe the experimental setup, including the datasets (Sec.~\ref{sec:exp_datasets}),  metrics (Sec.~\ref{sec:exp_metrics}), and the parametric values employed for provenance image filtering (Sec.~\ref{sec:exp_p1}) and provenance graph construction (Sec.~\ref{sec:exp_p2}).

\subsection{Datasets}
\label{sec:exp_datasets}
%In this paper, we are conducting experiments over three %different provenance datasets.
%They are detailed as it follows.

\subsubsection{NIST Dataset}
As a part of the \emph{Nimble Challenge 2017}~\cite{nist2017dataset}, NIST released a dataset specifically curated for the tasks of provenance image filtering and graph construction.
Named \emph{NC2017-Dev1-Beta4}, it contains 65 queries and 11,040 images that comprise samples related to the queries and distractors.
As a consequence, the dataset makes available a complete groundtruth that is composed of the 65 expected image ranks as well as the 65 expected provenance graphs related to each query.
The provenance graphs were manually created and include images resulting from a wide range of transformations, such as splicing, removal, cropping, scaling, rotation, translation and color correction.

Aiming to enlarge NC2017-Dev1-Beta4 towards a more realistic scenario, we extend its set of distractors by adding %1,038,684 
nearly one million images randomly sampled from the Nimble \emph{NC2017-Eval-Ver1} dataset~\cite{nist2017dataset}.
The NC2017-Eval-Ver1 dataset is the latest NIST evaluation set for measuring the performance of diverse image-manipulation detection tasks.
However, no complete provenance ground truth is available for this set, leading us to use NC2017-Dev1-Beta4 in conjunction with NC2017-Eval-Ver1.
As a result, we end up with what we call the \textit{NIST dataset}, which comprises the 65 provenance graphs from NC2017-Dev1-Beta4 and more than one million distractors from both datasets.

Following NIST suggestions in~\cite{nist2017dataset}, we perform both \emph{end-to-end} and \emph{oracle-filter} provenance analysis over the NIST dataset.
On the one hand, the end-to-end analysis includes performing the provenance image filtering task first, and then submitting the obtained image rank to the provenance graph construction step.
On the other hand, the oracle-filter analysis focuses on the provenance graph construction task; it assumes that a perfect image filtering solution is available.
Therefore, only the graph construction step is evaluated. %, using the groundtruth of the image filtering step as input.

\vspace{0.2cm}
\subsubsection{Professional Dataset}
Oliveira et. al~\cite{Oliveira_2016} introduced a multiple-parent phylogeny dataset, which comprises composite forgeries that always have two direct ancestors, namely the \emph{host} (which is used for defining the background of the composite) and the \emph{donor} (which they call \emph{alien} and that donates a local portion, such as an object or person, to define the foreground of the composite).
Each phylogeny case comprises 75 %high-resolution JPEG
images, of which three represent the composite, the host, and the donor, and the remaining 72 represent transformations (\textit{e.g.}, cropping, rotation, scale, and color transformations) over those three images.
As a consequence, each case is a provenance graph composed of three independent phylogeny trees (one for the host, one for the donor, and one for the composite) that are connected through the composite and its direct parents (the host and the donor, as expected).

Although our approach is not directly comparable to the one of Oliveira et al.~\cite{Oliveira_2016} (since they used different metrics and addressed a different problem of finding the correct original images --- the graph sources --- rather than the quality of the coverage of the complete provenance graph) we make use of their dataset for the reason of the composites being the work of a professional artist that tried to make the images as credible as possible.
Therefore, we are assessing the metrics defined in Sec.~\ref{sec:exp_metrics} and reporting the results over the 80 test cases found within the dataset.
In order to adapt it to our provenance graph building pipeline, however, we are choosing a random image inside the provenance graph as a query for each one of the 80 experimental cases.
Finally, we do not extend the professional dataset with distractors; hence we perform only oracle-filter analysis over it.

\vspace{0.2cm}
\subsubsection{Reddit Dataset}
\begin{figure}[t]
\centering
\includegraphics[width=8.5cm]{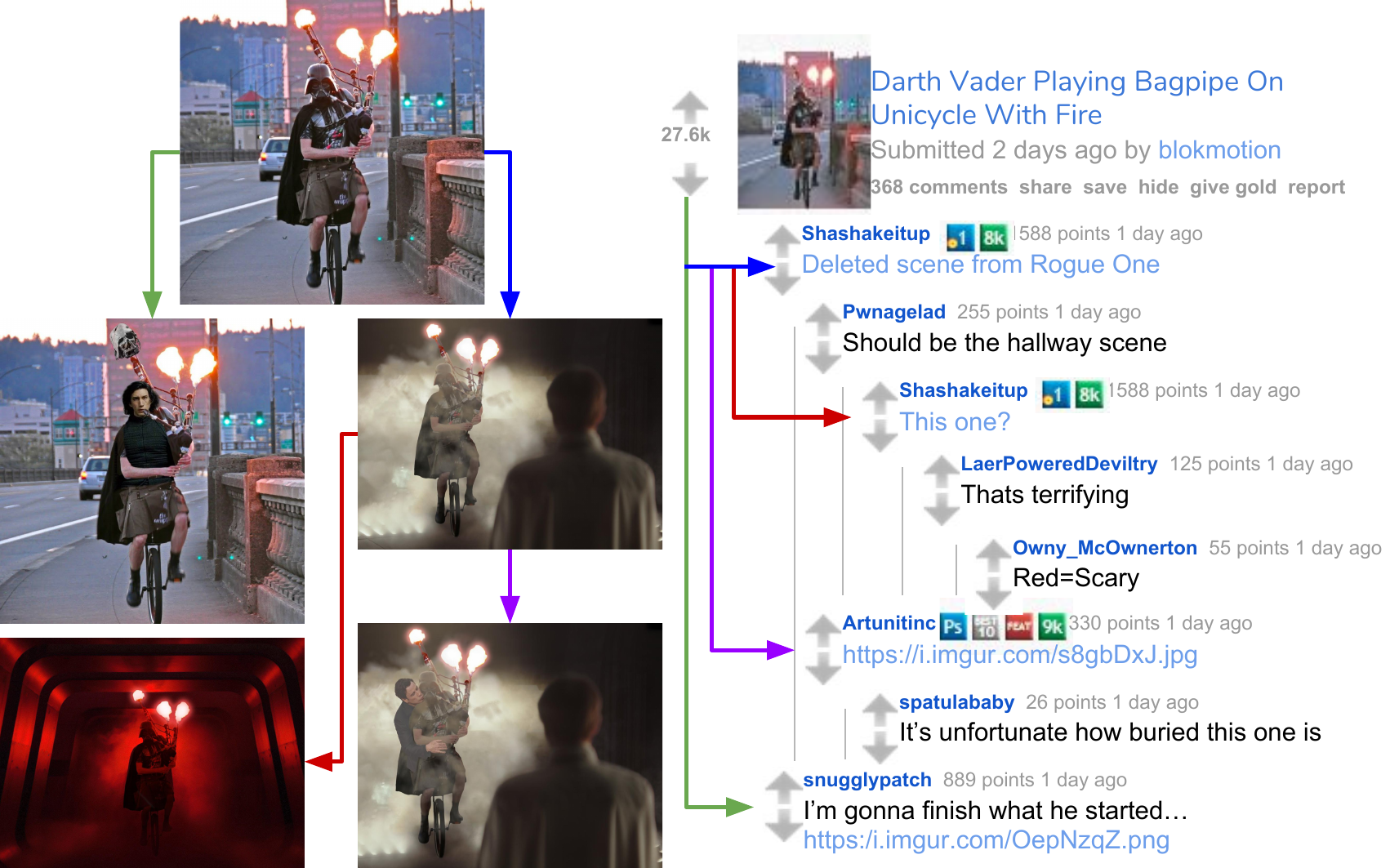}
\caption[]{A visualization of how provenance graphs are automatically inferred from a Reddit Photoshop battle instance.
The parent-child behavior of comments (right) can be leveraged to infer the structure of the ground truth provenance graph (left).
The colors of each comment correspond to their respective edge in the graph.} %\footnotemark.}
\label{fig:redditViz}
\end{figure}
%\footnotetext{\url{https://www.reddit.com/r/photoshopbattles/comments/76iqgc/psbattle_darth_vader_playing_bagpipe_on_unicycle/} (accessed on Oct. 31, 2017)}

To supplement the experimental data with even more realistic examples, we have collected a new provenance dataset from image content posted to the online Reddit community known as \emph{Photoshop battles}~\cite{reddit2017photoshopbattles}.
This community provides a medium for professional and amateur image manipulators to experiment with image doctoring in an environment of friendly competition.
Each ``battle'' begins with a single root image submitted by a user.
Subsequent users then post different modifications, usually humorous, of the image as comments to the original post.
Due to the competitive nature of the community, many image manipulations build off one another, as users try to outdo each other for comic effect.
This results in manipulation provenance trees with both wide and deep chains.
We use the underlying comment structure of these battles to automatically infer the ground truth provenance graph structure, as shown in Fig.~\ref{fig:redditViz}.

Because these images are real examples of incremental manipulations, the Reddit dataset accurately represents manipulations and operations performed on images in the wild.
%This is in contrast to all other provenance and phylogeny datasets, which contain only laboratory generated examples.
In total, the Reddit dataset contains 184 provenance graphs, which together sum up to 10,421 original and composite images.
It will be made available to the public upon the publication of this work.
Similar to the Professional dataset, we are not extending the Reddit dataset with distractors; we perform only oracle-filter analysis over it.

\subsection{Evaluation Metrics}
\label{sec:exp_metrics}

In this work, we adopt the metrics proposed by NIST in~\cite{nist2017dataset} for both the provenance image filtering and graph construction tasks.
In the case of provenance image filtering, we report (for each image query) the CBIR recall of the expected images at three particular cut-off ranks: $R@50$ (the recall considering the top-50 images of the retrieved image rank), $R@100$ (recall for the top-100 images), and $R@200$ (recall for the top-200 images).
Given that recall expresses the percentage of relevant images that are being effectively retrieved, the solution delivering higher recall is considered preferable.

In the case of provenance graph construction, we assess, for each provenance graph that is computed for each query, the $F_1$-measure (\textit{i.e.}, the harmonic mean of precision and recall) of the \emph{retrieved nodes} and of the \emph{retrieved edges} (called \emph{vertex overlap} ($VO$) and \emph{edge overlap} ($EO$), respectively).
Additionally, we report the \emph{vertex and edge overlap} ($VEO$), which is the $F_1$-measure of retrieving both nodes and edges, simultaneously~\cite{Papadimitriou_2010}.
The aim of using such metrics is to assess the overlap between the groundtruth and the constructed provenance graph.
The higher the values of $VO$, $EO$, and $VEO$, the better the quality of the solution.

Finally, in the particular case of $EO$ (and consequently $VEO$), we report the overlap both for directed edges (which are assumed to be the regular situation, and therefore kept for $EO$ and $VEO$), and for undirected edges (when an edge is considered to overlap another one if they connect analogous pairs of nodes, in spite of their orientations).
All aforementioned metrics are assessed through the NIST \emph{MediScore} tool~\cite{nist2017plan}.

\subsection{Filtering Setup}
\label{sec:exp_p1}
In all provenance filtering experiments, we either start describing the images with regular 64-dimensional SURF~\cite{Bay:CVIU:2008} interest points, or the distributed approach explained in Sec.~\ref{sec:prop_spread_kp} combined with the SURF detector (namely \emph{DSURF}).
For the regular SURF detector, depending on the experiment, we either extract the top-$2,000$ most responsive interest points (namely \emph{SURF2k}), or the top-$5,000$ most responsive ones (\emph{SURF5k}).
DSURF, in turn, is always described with 5,000 64-dimensional interest points, of which 2,500 regard the top-$2,500$ most responsive ones, and the remaining 2,500 are obtained avoiding overlap, as explained in~Sec.~\ref{sec:prop_spread_kp}.

For the sake of comparison, besides reporting results of the IVFADC system (explained in Sec.~\ref{sec:prop_indxing}), we also report results of the KD-Forest system discussed by Pinto~et~al.~\cite{pinto2017filtering} over the same set of images.
Because the work in~\cite{pinto2017filtering} is not easily scalable beyond 2,000 interest points, with respect to memory footprint, we combine it with SURF2k only (namely \emph{KDF-SURF2k}).

Focusing on the IVFADC approach, we provide combinations of it with all the available low-level descriptor approaches, hence obtaining \emph{IVFADC-SURF2k} (for comparison with KDF-SURF2k), \emph{IVFADC-SURF5k}, and \emph{IVFADC-DSURF}.
Regardless of the descriptors, we are always performing IVFADC with a codebook set size of 32 codes and sub-codebook set size of 96; both values were learned from preliminary experiments as revealing an acceptable trade-off between index building time and size, and final system recall. 
% TODO: Daniel: ask Joel about this codebook and sub-codebook thing. We need to add it to the method explanation (it seems to be missing, for now).
Finally, aiming at evaluating the impact of using iterative filtering (explained in Sec.~\ref{sec:prop_iterative_filtering}), we evaluate variations of the two most robust filtering solutions (namely IVFADC-SURF5k and IVFADC-DSURF) by adding iterative filtering (IF), hence obtaining the \emph{IVFADC-SURF5k-IF} and \emph{IVFADC-DSURF-IF} variations.
% TODO: Daniel: verify with Joel how many iterations we are employing in IF. We should report that here.
% I found it here: Second-tier search utilized a maximum of top 10 result and a minimum of the top 1 images from each first-tier query (chosen using RCMM, Section \ref{lab:filt:SecondTierSearch}.
All filtering methods are tested over the NIST dataset, for each one of its 65 queries.

% TODO: Daniel: describe the computational infrastructure used in the experiments.
% Good Joel's starting points:
% 1. It should be noted that these experiments were performed on a high-performance computing cluster utilizing the Andrew File System (AFS) [60].
% 2. All setups use 5000 descriptors per image and 8 TitanXP GPUs for building the index.
% Add CPU and GPU info (type, number of instances, amount of memory, etc.)

\subsection{Graph Construction Setup}
\label{sec:exp_p2}
As explained in Sec.~\ref{sec:prop:graph_building}, the graph construction task always starts with a given query and its respective rank of potentially related images.
For computing both the GCM-based and MI-based dissimilarity matrices (all explained in Sec.~\ref{sec:prop_dismat}), we either detect and match the top-5,000 most responsive SURF interest points per image, for each image pair, or the top-5,000 largest MSER regions per image, again for each image pair.
As a consequence, we have available four types of dissimilarity matrices, namely \emph{GCM-SURF} and \emph{GCM-MSER} (both symmetric) and \emph{MI-SURF} and \emph{MI-MSER} (both asymmetric).

The reason for choosing SURF and MSER is related to their potential complementarity: while SURF detects blobs of interest~\cite{Bay:CVIU:2008}, MSER detects the stable complex image regions that are tolerant to various perspective transformations~\cite{Matas_2004}.
Thus, the two methods end up delivering very different sets of interest points.
For extracting feature vectors from both SURF and MSER detected interest points, we compute the 64-dimensional SURF features proposed in~\cite{Bay:CVIU:2008}.
In the particular case of MSER, we compute the SURF features over the minimum enclosing circles that contain each one of the detected MSER image regions.
During the GCM feature matching, we match only interest points of the same type (\textit{i.e.}, we match SURF blobs with only SURF blobs, as well as MSER regions with only MSER regions).

In the end, we construct the provenance graphs from each one of the four types of dissimilarity matrices using either Kruskal's algorithm over the symmetric GCM-based instances (therefore obtaining undirected graphs), or the herein proposed clustered provenance graph expansion approach over the asymmetric MI-based instances (obtaining directed graphs). 
% TODO: Daniel: check with Aparna if we are using fusion of methods. If so, we need to explain fusion in the proposed method.

All graph construction methods are tested over the NIST (65 queries), Professional (80 queries), and Reddit (184 queries) datasets.
In the particular case of the NIST dataset, we report both end-to-end and oracle-filter analyses.
Regarding end-to-end analysis, we start with the best top-100 image ranks that were obtained in the former set of provenance image filtering experiments.
As expected, these ranks contain distractors, as well as miss some images related to the query that should be part of the final provenance graph.
With respect to the oracle-filter analysis, the ranks will only contain images related to the query.
%The same aspect will be observed in the case of the Professional and Reddit datasets, which do not count on distractors and, as a result, are submitted to oracle-filter analysis only.
% TODO: Daniel: describe the computational infrastructure used in the experiments.
% Add CPU info (type, number of cores, amount of memory, etc.)
              % Experimental Setup
% Daniel's notes: this is a new version of the results section, which reduces the amount of reported results and moves experimental details to the previous experimental setup section.
% Please refer to 05_results.tex for the original version.

\section{Results}
\label{sec:results}
In this section, we report the experimental results concerning the tasks of provenance image filtering (in Sec.~\ref{sec:res_1}) and of provenance graph construction (in Sec.~\ref{sec:res_2}).

\subsection{Image Filtering}
\label{sec:res_1}
Table~\ref{tab:filtering} contains the results of provenance image filtering over the 65 queries of the NIST dataset, following the setup detailed in Sec.~\ref{sec:exp_p1}.
The best solution is IVFADC-DSURF-IF, which reaches an R@50 value of 0.907, meaning that, if we use the respective top-50 rank as input to the provenance task, an average of 90.7\% of the images directly and indirectly related to the query will be available for graph construction.

As one might observe, the IVFADC-based solutions presented better recall values when compared to KDF-SURF2k, even when the same number of interest points was used for describing the images of the dataset.
That is the case, for instance, of the use of IVFADC-SURF2k, which provided an increase of approximately 17\% in R@50 over its KDF-based counterpart (KDF-SURF2k).
IVFADC makes use of CBIR state-of-the-art OPQ, which appears to be more effective than KD-trees for indexing image content.

In addition, the GPU-amenable scalability provided by IVFADC allowed us to increase the number of 64-dimensional SURF interest points from 2,000 to 5,000 features per image (reaching around five billion feature vectors for the entire dataset).
With more interest points, the dataset is better described, leading, for example, to an increase of nearly 23\% in R@50 for IVFADC-SURF2k over IVFADC-SURF5k.

\begin{table}[t] %*}[t]
\renewcommand{\arraystretch}{1.5}
\caption{Results of provenance image filtering over the NIST dataset.}
\centering
\begin{tabular}{R{2.3cm}C{1.5cm}C{1.5cm}C{1.5cm}} %C{2cm}C{2cm}C{2cm}}
    \hline
    Solution & R@50$^\dagger$ & R@100$^\dagger$ & R@200$^\dagger$ \\ %& Search time$^\dagger$ (min) & Index size (GB) & Indexing time (h) \\
    \hline
    KDF-SURF2k~\cite{pinto2017filtering} & 0.609 & 0.633 & 0.649 \\ %&      &      &    \\
    IVFADC-SURF2k    & 0.713 & 0.722 & 0.738 \\ %&      &      &    \\
    IVFADC-SURF5k    & 0.876 & 0.881 & 0.883 \\ %& 0.55 & 176.0 & 5 \\
    IVFADC-DSURF     & 0.882 & 0.895 & 0.899 \\ %& 0.54 & 152.8 & 6 \\
    IVFADC-SURF5k-IF & 0.895 & 0.901 & 0.919 \\ %& 2.53 & 176.0 & 5 \\
    \textbf{IVFADC-DSURF-IF}  & \textbf{0.907} & \textbf{0.912} & \textbf{0.923} \\ %& \textbf{2.20} & \textbf{152.8} & \textbf{6} \\
    \hline
\end{tabular}
\vspace{0.1cm}\\
$\dagger$: We report the average values on the provided 65 queries.\\
In bold, the solution with highest recall values.
\label{tab:filtering}
\end{table} %*}
% TODO: Daniel: ask Joel about R@100 < R@50 (line 2). It is probably not correct.
% TODO: Daniel: add standard deviations to these results.

The use of DSURF also increased the recall values.
Its application was responsible for an improvement of almost 7\% in R@50, when we compare IVFADC-SURF5k and IVFADC-DSURF, at the expense of adding one more hour %(from five to six hours)
to the time required to construct the index for the entire dataset.
This extra hour is related to the additional step of avoiding interest point overlaps, which is part of the DSURF detection solution.

Finally, the use of IF made the recall values approach 0.9, even considering R@50.
For example, the use of IVFADC-DSURF-IF yielded an improvement of nearly 3\% in R@50 over IVFADC-DSURF.
That happened, however, at the expense of a significant increase of search time, due to the iterative re-querying nature of IF; IVFADC-DSURF-IF requires four times longer than IVFADC-DSURF.
However, in certain scenarios where time is not a constraint, the increase of 3\% in recall may justify the deployment of such approach.

\subsection{Graph Construction}
\label{sec:res_2}
We organize the results of graph construction according to the adopted dataset (either NIST, Professional, or Reddit).
Table~\ref{tab:nist} shows the performance of the proposed approach over the NIST dataset.
%As previously explained in Sec.~\ref{sec:exp_p2},
Results are grouped into end-to-end and oracle-filter analysis.
In the particular case of end-to-end analysis, top-100 rank lists were obtained with IVFADC-DSURF-IF filtering, the best approach reported in Table~\ref{tab:filtering}.
As a consequence, the respective provenance graphs are built, on average, without almost $9\%$ of the image nodes, which are not retrieved in the filtering step ($R@100 = 0.912$, in the case of IVFADC-DSURF-IF).
Oracle-filter analysis, in turn, starts from a perfect rank of images, containing all and only the graph image nodes.
That explains the higher values of VO in such group, at the expense of reducing EO.
The reduction of EO is explained by the availability of more related images in the step of graph construction, which increases the number of possible edges and misconnections.
It means that the present solutions are good at removing distractors, but there is still room to improve the effective connection of sharing-content images.
The best end-to-end solution is MI-SURF, retrieving, on average, directed provenance graphs with $0.613$ ground truth-graph coverage (VEO).
The best oracle-filter solution, in turn, is GCM-SURF, with $0.609$ undirected graph coverage.
%Nonetheless, GCM solutions deliver only undirected graphs (since they use Kruskal's algorithm, as explained in Sec.~\ref{sec:exp_p2}), leaving the problem of determining the edge directions open.

\begin{table}[!htbp]
\renewcommand{\arraystretch}{1.5}
\caption{Results of provenance graph construction over the NIST dataset.
We report the average values on the provided 65 queries.}
\centering
\begin{tabular}{C{1.3cm}R{1.8cm}C{1.1cm}C{1.1cm}C{1.1cm}}
    \hline
    & Solution & VO & EO & VEO\\
    \hline
    \multirow{4}{1.3cm}{End-to-end analysis} & GCM-SURF~\cite{bharati2017uphy} & 0.638 & 0.429$^\dagger$ & 0.537$^\dagger$ \\
                                             & GCM-MSER & 0.257 & 0.140$^\dagger$ & 0.199$^\dagger$ \\
                                             & \textbf{MI-SURF}  & \textbf{0.853} & \textbf{0.353} & \textbf{0.613} \\
                                             & MI-MSER  & 0.835 & 0.312 & 0.585 \\
    \hline
    \multirow{4}{1.3cm}{Oracle-filter analysis} & \textbf{GCM-SURF}~\cite{bharati2017uphy} & \textbf{0.933} & \textbf{0.256}$^\dagger$ & \textbf{0.609}$^\dagger$ \\
                                                & GCM-MSER & 0.902 & 0.239$^\dagger$ & 0.585$^\dagger$ \\
                                                & MI-SURF  & 0.931 & 0.124 & 0.546 \\
                                                & MI-MSER  & 0.892 & 0.123 & 0.525 \\
    \hline
\end{tabular}
\vspace{0.1cm}\\
$\dagger$: Values for undirected edges. In bold, the solutions with the best VEO.
\label{tab:nist}
\vspace{0.1cm}
\end{table}
% TODO: Daniel: add standard deviations

\begin{table}[!htbp]
\renewcommand{\arraystretch}{1.5}
\caption{Results of provenance graph construction over the Professional dataset.
We report the average values on the 80 queries belonging to the test set.}
\centering
\begin{tabular}{R{1.9cm}C{1.5cm}C{1.5cm}C{1.5cm}}
    \hline
    Solution & VO & EO & VEO\\
    \hline
    \textbf{GCM-SURF}~\cite{bharati2017uphy} & \textbf{0.985} & \textbf{0.218}$^\dagger$ & \textbf{0.604}$^\dagger$ \\
    GCM-MSER & 0.663 & 0.087$^\dagger$ & 0.377$^\dagger$ \\
    MI-SURF  & 0.975 & 0.102 & 0.541 \\
    MI-MSER  & 0.604 & 0.043 & 0.326 \\
    \hline
\end{tabular}
\vspace{0.1cm}\\
$\dagger$: Values for undirected edges. In bold, the solution with the best VEO.
\label{tab:prof}
\vspace{0.1cm}
\end{table}
% TODO: Daniel: add standard deviations

\begin{table}[!htbp]
\renewcommand{\arraystretch}{1.5}
\caption{Results of provenance graph construction over the Reddit dataset.
We report the average values on the provided 100 queries.}
\centering
\begin{tabular}{R{1.9cm}C{1.5cm}C{1.5cm}C{1.5cm}}
    \hline
    Solution & VO & EO & VEO\\
    \hline
    GCM-SURF~\cite{bharati2017uphy} & 0.884 & 0.156$^\dagger$ & 0.523$^\dagger$ \\
    \textbf{GCM-MSER} & \textbf{0.924} & \textbf{0.121}$^\dagger$ & \textbf{0.526}$^\dagger$ \\
    MI-SURF  & 0.757 & 0.037 & 0.401 \\
    MI-MSER  & 0.509 & 0.027 & 0.271 \\
    \hline
\end{tabular}
\vspace{0.1cm}\\
$\dagger$: Values for undirected edges. In bold, the solution with the best VEO.
\label{tab:reddit}
\end{table}
% TODO: Daniel: add standard deviations

In Table \ref{tab:prof}, we present results of the proposed approaches on the Professional dataset.
In comparison to the NIST dataset, the same solutions recognize fewer of the correct provenance graph edges.
This happens due to the larger 75-node provenance graphs, which contain a number of near duplicates that were created through reversible operations.
As a consequence, altered image nodes can be achieved using different sequences of image transformations, leading to ambiguous dissimilarity values, and multiple plausible paths, within the provenance graph.
The methods herein discussed are solely based on image content and do not consider any extra information, thus operating with data from only the pixel domain.
Indeed, in previous image phylogeny work reporting results on the Professional dataset, the solutions made use of data from the JPEG compression tables of the images.
We speculate that if information regarding the compression factor is included in the present approaches, some confusion regarding the edges can be eliminated.
That would not impact the NIST dataset, though, since only a small fraction of its images are available in JPEG format.
Here, the best solution is GCM-SURF, retrieving, on average, undirected provenance graphs with $0.604$  ground truth-graph coverage (VEO).

%In table \ref{tab:prof}, we present results of the proposed approaches on the Professional dataset. In comparison with the performance on the NIST dataset, the methods recognize fewer of the correct provenance graph edges. The professional dataset consists of large provenance cases with 75 images considered for each provenance graph. Also, the dataset contains a number of near-duplicate cases that have been created through operations that are reversible and similar effects can be achieved using different sequences of transformations. This leads to close dissimilarity values for these cases within provenance making each one of them a plausible path within the provenance graph. The approach presented here is solely based on image content and does not consider any extra information thus leaving the system just with information from the pixel domain, which is not comprehensive. In previous work on Multimedia Phylogeny that reports results on this dataset, the approaches utilize the information from the compression tables of the images since they are JPEGs. We speculate that if information regarding the compression factor is included in the approach, some confusion regarding the edges can be eliminated. This will mean that images that cannot provide such information will not be considered when establishing provenance, making our methods much less generalizable.

Table~\ref{tab:reddit} reports results on the Reddit dataset.
As one might observe, this dataset is the most challenging one, with low directed edge coverage (namely $EO$, in the case of MI-SURF and MI-MSER solutions).
Since its whimsical content is the product of a diverse community, the Reddit dataset presents realistic, yet frustratingly complex, cases.
As a consequence, it is not uncommon to find among the 184 collected provenance graphs suppressed ancestral images, as well as descendant images whose parental connections are defined by very particular and contextual semantic reasons (for instance, an arbitrary person resembling another in the parent image), than by strictly shared visual content.
That ends up impacting our results.
The best solution is GCM-MSER, which retrieves, on average, undirected provenance graphs with $0.526$  ground truth-graph coverage (VEO).

%Another example of the evaluation of the proposed approach in the real-world is the testing of the method in verifying provenance graphs on Reddit. Even though the dataset can have missing images because the source of alien objects used in creating compositions might not be present on Reddit, this is still considered an oracle evaluation because even the ground truth does not contain those images. All the images that are part of the ground truth are considered for provenance graph construction. 

% Daniel: I added the following content to conclusion.
%Upon scrutinizing the results from three differently sourced datasets having different types of transformations as part of the relationships between images, it can be observed that the proposed approach performs decently well in connecting the correct set of images (~0.8 vertex overlap) but fails at pointing out the correct directions ($<$0.5 edge overlap). Directions within edges are dependent on whether the transformation is reversible or can be inferred from pixel information. In this attempt to perform provenance analysis, we found that although image content is the most reliable source of information connecting related images, other external information may be required to supplement the knowledge obtained from pixels. This external information can be obtained from file metadata, object detectors and compression factors where available. 
               % Results
\section{Conclusions}
\label{sec:conc}

The determination of image provenance is a difficult task to solve. The complexity increases significantly when considering an end-to-end, fully-automatic provenance pipeline that performs at scale. This is the first work, to our knowledge, to have proposed such a technique, and we consider these experiments an important demonstration of the feasibility of large-scale provenance systems.

Our pipeline included an image indexing scheme that utilizes a novel iterative filtering and distributed interest point selection to provide results that outperform the current state-of-the-art found in \cite{pinto2017filtering}. %, while simultaneously improving index computation speeds significantly.
We also proposed methods for provenance graph building that improve upon the methods of previous work in the field, and  provided a novel clustering algorithm for further graph improvement.

To analyze these methods, we utilized the NIST Nimble Challenge~\cite{Nimble_2017} and the multiple-parent phylogeny Professional dataset~\cite{Oliveira_2016} to generate detailed performance results. Beyond utilizing these datasets, we committed to real-world provenance analysis by building our own dataset from Reddit~\cite{reddit2017photoshopbattles}, consisting of unique manipulation scenarios that were generated in an unconstrained environment. This is the first work of its kind to analyze fully in-the-wild provenance cases. 

Upon scrutinizing the results from the three differently sourced datasets, we observed that the proposed approaches perform decently well in connecting the correct set of images (with reported vertex overlaps of nearly 0.8), but still struggle when inferring edge directions --- a result that highlights the difficulty of this problem. Directed  edges are dependent on whether the transformations are reversible or can be inferred from pixel information. In this attempt to perform provenance analysis, we found that although image content is the most reliable source of information connecting related images, other external information may be required to supplement the knowledge obtained from pixels. This external information can be obtained from file metadata, object detectors and compression factors, whenever available.
 
%One major difficulty yet to be overcome in the case of provenance filtering is the detection and retrieval of small or non-salient alien objects within a composite image. While Iterative Filtering has been shown to help improve performance in these cases, we still find that the majority of images not correctly retrieved correspond to these small alien objects. We plan to explore the benefits of a new interest point selection and description algorithm, possibly utilizing deep local features and attention models, to replace SURF as the primary image characterization mechanism for filtering.

%Additionally, this work does not currently utilize previous work found in the Blind Digital Image Forensics (BDIF) field. Significant improvements in region localization, provenance edge calculation, and even edge directionality estimation could be realized by using systems already created in the BDIF field. We plan to explore the benefits of integrating splicing and copy-move detectors, along with noise \cite{}, Photo Response Non-Uniformity (PRNU) \cite{fridrich2009digital}, and Color Filter Array (CFA) models into our pipeline for detecting image inconsistencies and building higher accuracy dissimilarity matrices.

Work in this field is far from complete. The problem of unconstrained, fully-automatic image provenance analysis is not solved.  For instance, this work does not currently utilize previous work found in the Blind Digital Image Forensics (BDIF) field. Significant improvements in region localization, provenance edge calculation, and even edge direction estimation could be performed by using systems already created in the BDIF field. We plan to explore the benefits of integrating splicing and copy-move detectors, along with Photo Response Non-Uniformity (PRNU), %\cite{fridrich2009digital},
and Color Filter Array (CFA) models into our pipeline for detecting image inconsistencies and building higher accuracy dissimilarity matrices.

While this work is a significant first step, we hope to spur others on to further investigate fully-automatic image forensics systems. As the landscapes of social and journalistic media change, so must the field of image forensics adapt with them. News stories, cultural trends, and social sentiments flow at a fast pace, often fueled by unchecked viral images and videos. There is a pressing need to find new solutions and approaches to combat forgery and misinformation. Further, the dual-use nature of such systems makes them useful for other applications, such as cultural analytics, where image provenance can be a primary object of study. We encourage researchers to think broadly when it comes to image provenance analysis.
\section*{Acknowledgment}
This material is based on research sponsored by DARPA and Air Force Research Laboratory (AFRL) under agreement number FA8750-16-2-0173.
Hardware support was generously provided by the NVIDIA Corporation.
We also thank the financial support of FAPESP (Grant 2017/12646-3, D\'ej\`aVu Project), CAPES (DeepEyes Grant) and CNPq (Grant 304472/2015-8).

\ifCLASSOPTIONcaptionsoff
  \newpage
\fi

\bibliographystyle{IEEEtran}
\bibliography{bibliography}

% \begin{IEEEbiography}[{\includegraphics[width=1in,height=1.25in,clip,keepaspectratio]{picture}}]{John Doe}
% \blindtext
% \end{IEEEbiography}


%\vfill


\end{document}